# Genetic Algorithm for Solving Simple Mathematical Equality Problem

Denny Hermawanto

Indonesian Institute of Sciences (LIPI), INDONESIA

Mail: denny.hermawanto@gmail.com

Abstract

This paper explains genetic algorithm for novice in this field. Basic philosophy of genetic algorithm and its flowchart are described. Step by step numerical computation of genetic algorithm for solving simple mathematical equality problem will be briefly explained.

**Basic Philosophy**

Genetic algorithm developed by Goldberg was inspired by Darwin's theory of evolution which states that the survival of an organism is affected by rule "the strongest species that survives". Darwin also stated that the survival of an organism can be maintained through the process of reproduction, crossover and mutation. Darwin's concept of evolution is then adapted to computational algorithm to find solution to a problem called objective function in natural fashion. A solution generated by genetic algorithm is called a chromosome, while collection of chromosome is referred as a population. A chromosome is composed from genes and its value can be either numerical, binary, symbols or characters depending on the problem want to be solved. These chromosomes will undergo a process called fitness function to measure the suitability of solution generated by GA with problem. Some chromosomes in population will mate through process called crossover thus producing new chromosomes named offspring which its genes composition are the combination of their parent. In a generation, a few chromosomes will also mutation in their gene. The number of chromosomes which will undergo crossover and mutation is controlled by crossover rate and mutation rate value. Chromosome in the population that will maintain for the next generation will be selected based on Darwinian evolution rule, the chromosome which has higher fitness value will have greater probability of being selected again in the next generation. After several generations, the chromosome value will converges to a certain value which is the best solution for the problem.

**The Algorithm**

In the genetic algorithm process is as follows [1]:
Step 1. Determine the number of chromosomes, generation, and mutation rate and crossover rate value
Step 2. Generate chromosome-chromosome number of the population, and the initialization value of the genes chromosome-chromosome with a random value

Step 3. Process steps 4-7 until the number of generations is met
Step 4. Evaluation of fitness value of chromosomes by calculating objective function
Step 5. Chromosomes selection
Step 6. Crossover
Step 7. Mutation
Step 8. Solution (Best Chromosomes)

The flowchart of algorithm can be seen in Figure 1

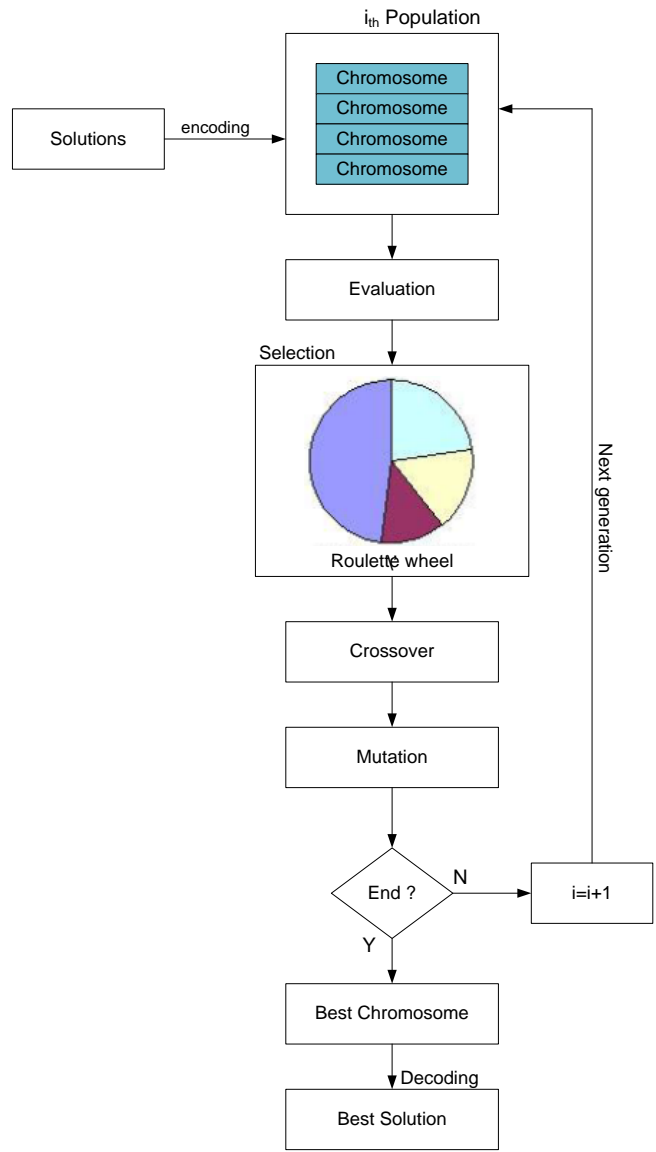

Figure 1. Genetic algorithm flowchart

**Numerical Example**

Here are examples of applications that use genetic algorithms to solve the problem of combination. Suppose there is equality **a** + 2**b** + 3**c** + 4**d** = 30, genetic algorithm will be used to find the value of **a**, **b**, **c**, and **d** that satisfy the above equation. First we should formulate the objective function, for this problem the objective is minimizing the value of function **f(x)** where **f(x)** = ((**a** + 2**b** + 3**c** + 4**d**) - 30). Since there are four variables in the equation, namely **a**, **b**, **c**, and **d**, we can compose the chromosome as follow: To speed up the computation, we can restrict that the values of variables **a**, **b**, **c**, and **d** are integers between 0 and 30.

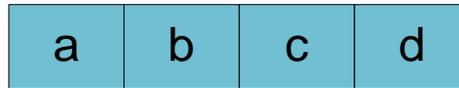

**Step 1. Initialization**
For example we define the number of chromosomes in population are 6, then we generate random value of gene **a**, **b**, **c**, **d** for 6 chromosomes
**Chromosome**[1] = [a;b;c;d] = [12;05;23;08]
**Chromosome**[2] = [a;b;c;d] = [02;21;18;03]
**Chromosome**[3] = [a;b;c;d] = [10;04;13;14]
**Chromosome**[4] = [a;b;c;d] = [20;01;10;06]
**Chromosome**[5] = [a;b;c;d] = [01;04;13;19]
**Chromosome**[6] = [a;b;c;d] = [20;05;17;01]

**Step 2. Evaluation**
We compute the objective function value for each chromosome produced in initialization step:
**F_obj**[1] = Abs(( 12 + 2*05 + 3*23 + 4*08 ) - 30)
 = Abs((12 + 10 + 69 + 32 ) - 30)
 = Abs(123 - 30)
 = 93
**F_obj**[2] = Abs((02 + 2*21 + 3*18 + 4*03) - 30)
 = Abs((02 + 42 + 54 + 12) - 30)
 = Abs(110 - 30)
 = 80
**F_obj**[3] = Abs((10 + 2*04 + 3*13 + 4*14) - 30)
 = Abs((10 + 08 + 39 + 56) - 30)
 = Abs(113 - 30)
 = 83
**F_obj**[4] = Abs((20 + 2*01 + 3*10 + 4*06) - 30)
 = Abs((20 + 02 + 30 + 24) - 30)
 = Abs(76 - 30)
 = 46
**F_obj**[5] = Abs((01 + 2*04 + 3*13 + 4*19) - 30)
 = Abs((01 + 08 + 39 + 76) - 30)
 = Abs(124 - 30)
 = 94
**F_obj**[6] = Abs((20 + 2*05 + 3*17 + 4*01) - 30)
 = Abs((20 + 10 + 51 + 04) - 30)

    = Abs(85 - 30)
    = 55

**Step 3. Selection**
1. The fittest chromosomes have higher probability to be selected for the next generation. To compute fitness probability we must compute the fitness of each chromosome. To avoid divide by zero problem, the value of **F_obj** is added by 1.

  **Fitness**[1] = 1 / (1+**F_obj**[1])
    = 1 / 94
    = 0.0106
  **Fitness**[2] = 1 / (1+**F_obj**[2])
    = 1 / 81
    = 0.0123
  **Fitness**[3] = 1 / (1+**F_obj**[3])
    = 1 / 84
    = 0.0119
  **Fitness**[4] = 1 / (1+**F_obj**[4])
    = 1 / 47
    = 0.0213
  **Fitness**[5] = 1 / (1+**F_obj**[5])
    = 1 / 95
    = 0.0105
  **Fitness**[6] = 1 / (1+**F_obj**[6])
    = 1 / 56
    = 0.0179

**Total** = 0.0106 + 0.0123 + 0.0119 + 0.0213 + 0.0105 + 0.0179
  = 0.0845

The probability for each chromosomes is formulated by: **P**[i] = **Fitness**[i] / **Total**
**P**[1] = 0.0106 / 0.0845
  = 0.1254
**P**[2] = 0.0123 / 0.0845
  = 0.1456
**P**[3] = 0.0119 / 0.0845
  = 0.1408
**P**[4] = 0.0213 / 0.0845
  = 0.2521
**P**[5] = 0.0105 / 0.0845
  = 0.1243
**P**[6] = 0.0179 / 0.0845
  = 0.2118

From the probabilities above we can see that Chromosome 4 that has the highest fitness, this chromosome has highest probability to be selected for next generation chromosomes. For the

selection process we use roulette wheel, for that we should compute the cumulative probability values:

**C**[1] = 0.1254
**C**[2] = 0.1254 + 0.1456
= 0.2710
**C**[3] = 0.1254 + 0.1456 + 0.1408
= 0.4118
**C**[4] = 0.1254 + 0.1456 + 0.1408 + 0.2521
= 0.6639
**C**[5] = 0.1254 + 0.1456 + 0.1408 + 0.2521 + 0.1243
= 0.7882
**C**[6] = 0.1254 + 0.1456 + 0.1408 + 0.2521 + 0.1243 + 0.2118
= 1.0

Having calculated the cumulative probability of selection process using roulette-wheel can be done. The process is to generate random number **R** in the range 0-1 as follows.

**R**[1] = 0.201
**R**[2] = 0.284
**R**[3] = 0.099
**R**[4] = 0.822
**R**[5] = 0.398
**R**[6] = 0.501

If random number **R**[1] is greater than **C**[1] and smaller than **C**[2] then select **Chromosome**[2] as a chromosome in the new population for next generation:

**NewChromosome**[1] = **Chromosome**[2]
**NewChromosome**[2] = **Chromosome**[3]
**NewChromosome**[3] = **Chromosome**[1]
**NewChromosome**[4] = **Chromosome**[6]
**NewChromosome**[5] = **Chromosome**[3]
**NewChromosome**[6] = **Chromosome**[4]

Chromosomes in the population thus became:
**Chromosome**[1] = [02;21;18;03]
**Chromosome**[2] = [10;04;13;14]
**Chromosome**[3] = [12;05;23;08]
**Chromosome**[4] = [20;05;17;01]
**Chromosome**[5] = [10;04;13;14]
**Chromosome**[6] = [20;01;10;06]

In this example, we use one-cut point, i.e. randomly select a position in the parent chromosome then exchanging sub-chromosome. Parent chromosome which will mate is randomly selected and the number of mate Chromosomes is controlled using **crossover_rate** (**ρc**) parameters. Pseudo-code for the crossover process is as follows:

```
begin
   k← 0;
   while(k<population) do
   R[k] = random(0-1);
   if(R[k]< ρc) then
      select Chromosome[k] as parent;
   end;
   k = k + 1;
   end;
end;
```

Chromosome **k** will be selected as a parent if **R**[k]<ρc. Suppose we set that the crossover rate is 25%, then Chromosome number **k** will be selected for crossover if random generated value for Chromosome **k** below 0.25. The process is as follows: First we generate a random number **R** as the number of population.
**R**[1] = 0.191
**R**[2] = 0.259
**R**[3] = 0.760
**R**[4] = 0.006
**R**[5] = 0.159
**R**[6] = 0.340

For random number **R** above, parents are **Chromosome**[1], **Chromosome**[4] and **Chromosome**[5] will be selected for crossover.
**Chromosome**[1] >< **Chromosome**[4]
**Chromosome**[4] >< **Chromosome**[5]
**Chromosome**[5] >< **Chromosome**[1]

After chromosome selection, the next process is determining the position of the crossover point. This is done by generating random numbers between 1 to (length of Chromosome − 1). In this case, generated random numbers should be between 1 and 3. After we get the crossover point, parents Chromosome will be cut at crossover point and its gens will be interchanged. For example we generated 3 random number and we get:
**C**[1] = 1
**C**[2] = 1
**C**[3] = 2

Then for first crossover, second crossover and third crossover, parent's gens will be cut at gen number 1, gen number 1 and gen number 3 respectively, e.g.
**Chromosome**[1] = **Chromosome**[1] >< **Chromosome**[4]
         = [02;21;18;03] >< [20;05;17;01]
         = [02;05;17;01]
**Chromosome**[4] = **Chromosome**[4] >< **Chromosome**[5]
         = [20;05;17;01] >< [10;04;13;14]

= [20;04;13;14]
**Chromosome**[5] = **Chromosome**[5] >< **Chromosome**[1]
= [10;04;13;14] >< [02;21;18;03]
= [10;04;18;03]

Thus Chromosome population after experiencing a crossover process:
**Chromosome**[1] = [02;05;17;01]
**Chromosome**[2] = [10;04;13;14]
**Chromosome**[3] = [12;05;23;08]
**Chromosome**[4] = [20;04;13;14]
**Chromosome**[5] = [10;04;18;03]
**Chromosome**[6] = [20;01;10;06]

**Step 5. Mutation**
Number of chromosomes that have mutations in a population is determined by the **mutation_rate** parameter. Mutation process is done by replacing the gen at random position with a new value. The process is as follows. First we must calculate the total length of gen in the population. In this case the total length of gen is **total_gen** = **number_of_gen_in_Chromosome** * **number of population**
= 4 * 6
= 24

Mutation process is done by generating a random integer between 1 and total_gen (1 to 24). If generated random number is smaller than mutation_rate($\rho$m) variable then marked the position of gen in chromosomes. Suppose we define $\rho$m 10%, it is expected that 10% (0.1) of total_gen in the population that will be mutated:
number of mutations = 0.1 * 24
= 2.4
$\approx$ 2

Suppose generation of random number yield 12 and 18 then the chromosome which have mutation are Chromosome number 3 gen number 4 and Chromosome 5 gen number 2. The value of mutated gens at mutation point is replaced by random number between 0-30. Suppose generated random number are 2 and 5 then Chromosome composition after mutation are:
**Chromosome**[1] = [02;05;17;01]
**Chromosome**[2] = [10;04;13;14]
**Chromosome**[3] = [12;05;23;**02**]
**Chromosome**[4] = [20;04;13;14]
**Chromosome**[5] = [10;**05**;18;03]
**Chromosome**[6] = [20;01;10;06]

Finishing mutation process then we have one iteration or one generation of the genetic algorithm. We can now evaluate the objective function after one generation:

**Chromosome**[1] = [02;05;17;01]  
**F_obj**[1] = Abs(( 02 + 2*05 + 3*17 + 4*01 ) - 30)  
       = Abs((2 + 10 + 51 + 4 ) - 30)  
       = Abs(67 - 30)  
       = 37  
**Chromosome**[2] = [10;04;13;14]  
**F_obj**[2] = Abs(( 10 + 2*04 + 3*13 + 4*14 ) - 30)  
       = Abs((10 + 8 + 33 + 56 ) - 30)  
       = Abs(107 - 30)  
       = 77  
**Chromosome**[3] = [12;05;23;02]  
**F_obj**[3] = Abs(( 12 + 2*05 + 3*23 + 4*02 ) - 30)  
       = Abs((12 + 10 + 69 + 8 ) - 30)  
       = Abs(87 - 30)  
       = 47  
**Chromosome**[4] = [20;04;13;14]  
**F_obj**[4] = Abs(( 20 + 2*04 + 3*13 + 4*14 ) - 30)  
       = Abs((20 + 8 + 39 + 56 ) - 30)  
       = Abs(123 - 30)  
       = 93  
**Chromosome**[5] = [10;05;18;03]  
**F_obj**[5] = Abs(( 10 + 2*05 + 3*18 + 4*03 ) - 30)  
       = Abs((10 + 10 + 54 + 12 ) - 30)  
       = Abs(86 - 30)  
       = 56  
**Chromosome**[6] = [20;01;10;06]  
**F_obj**[6] = Abs(( 20 + 2*01 + 3*10 + 4*06 ) - 30)  
       = Abs((20 + 2 + 30 + 24 ) - 30)  
       = Abs(76 - 30)  
       = 46  

From the evaluation of new Chromosome we can see that the objective function is decreasing, this means that we have better Chromosome or solution compared with previous Chromosome generation. New Chromosomes for next iteration are:

**Chromosome**[1] = [02;05;17;01]  
**Chromosome**[2] = [10;04;13;14]  
**Chromosome**[3] = [12;05;23;02]  
**Chromosome**[4] = [20;04;13;14]  
**Chromosome**[5] = [10;05;18;03]  
**Chromosome**[6] = [20;01;10;06]  

These new Chromosomes will undergo the same process as the previous generation of Chromosomes such as evaluation, selection, crossover and mutation and at the end it produce new generation of Chromosome for the next iteration. This process will be repeated until a predetermined number of generations. For this example, after running 50 generations, best chromosome is obtained:

Chromosome = [07; 05; 03; 01]

This means that: **a** = 7, **b** = 5, **c** = 3, **d** = 1

If we use the number in the problem equation:

**a** + 2**b** + 3**c** + 4**d** = 30

7 + (2 * 5) + (3 * 3) + (4 * 1) = 30

We can see that the value of variable **a**, **b**, **c** and **d** generated by genetic algorithm can satisfy that equality.